\documentclass{isprs} % isprs class modified 23-04-2019 (Dennis Wittich)
\usepackage{subfigure}
\usepackage{setspace}
\usepackage{geometry} % added 27-02-2014 Markus Englich
\usepackage{epstopdf}
\usepackage[labelsep=period]{caption}  % added 14-04-2016 Markus Englich - Recommendation by Sebastian Brocks
\usepackage[british]{babel} 
\usepackage[hang]{footmisc}
\usepackage{url}

\usepackage{subfig}
\usepackage{booktabs}
\usepackage{multirow}
\usepackage{xcolor}
\begin{document}

\title{Space2Ground 2.0: A Multi-Source Dataset and Framework for Agricultural Monitoring through Fusion of Street-Level and Satellite Imagery}
\date{}

% KAO: Remove extra spacing

% Anonymous submissions, authors' names should not be visible
\author{
 Iason Tsardanidis\textsuperscript{1}, Alkiviadis Koukos\textsuperscript{2}, George Choumos\textsuperscript{1}, Vasileios Sitokonstantinou\textsuperscript{3}, Charalampos Kontoes\textsuperscript{1} }
% \author{***** (for review, names must be rendered anonymous)}

% KAO: Remove extra newline
% Anonymous submissions, authors' affiliations should not be visible
\address{
	\textsuperscript{1 }Operational Unit BEYOND Centre, IAASARS, National Observatory of Athens, Athens, Greece \\- (j.tsardanidis, g.choumos, kontoes)@noa.gr\\
	\textsuperscript{2 }DHI Water \& Environment, Hørsholm, Denmark - akou@dhigroup.com\\
	\textsuperscript{3 }Artificial Intelligence Group, Wageningen University \& Research, The Netherlands – vassilis.sitokonstantinou@wur.nl\\
	% \textsuperscript{4 }Univ. Gustave Eiffel, IGN-ENSG, LaSTIG – Saint-Mandé, France – clement.mallet@ign.fr\\
	% \textsuperscript{5 }Institute of Photogrammetry and GeoInformation, Leibniz Universit\"at Hannover, Germany - heipke@ipi.uni-hannover.de\\
}
% \address{**** (for review, affiliations must be rendered anonymous)}

% If the corresponding author is NOT the final author, always add a % space before the subsequent comma, i.e.
% first author name\textsuperscript{a,}\thanks{Corresponding author} , % second author name \textsuperscript{b}, etc.
% thanks to Niclas Borlin 05-05-2016
% information on the corresponding author should not be used any longer and has been commented out
% C. Heipke, Jan 03,2024

% the use of the information of commissions and working groups should not be used any longer and has been commented out
% C. Heipke, Sept. 20,2022
%\commission{XX, }{YY} %This field is optional. If filled, XX and YY should be replaced by adequate numbers. See https://www2.isprs.org/commissions/
%\workinggroup{XX/YY} %This field is optional.
%\icwg{}   %This field is optional.

% KAO: Use times symbol
\abstract{
Accurate and scalable parcel-level agricultural monitoring remains challenging because satellite Earth Observation alone provides only an overhead perspective of agricultural parcels, while optical observations are further affected by cloud-induced temporal gaps.
This paper presents Space2Ground 2.0, a multi-source framework integrating Sentinel-1 SAR and Sentinel-2 multispectral time series with geo-tagged street-level imagery acquired using vehicle-mounted cameras and shared through the Mapillary platform. A largely automated processing pipeline performs semantic filtering, image quality assessment, viewpoint-based parcel association, and dataset refinement, transforming large volumes of crowdsourced imagery into parcel-linked, analysis-ready data.
Applied over Cyprus during the 2022 growing season, the pipeline produced a curated dataset of 46,050 annotated street-level images, selected from an initial collection exceeding 900,000 images and linked with satellite information for 8,581 agricultural parcels. The practical value of the dataset was assessed through parcel-level crop classification experiments using both single- and multi-source observations. 
% The best late-fusion configuration achieved an overall accuracy of 84.12\% and a macro-F1-score of 82.78\%, improving overall accuracy by 5.22 \% compared with the strongest satellite-only model.
The results demonstrate that street-level imagery provides complementary fine-scale visual information that enhances classification when integrated with satellite time series. Overall, Space2Ground 2.0 provides an openly available benchmark dataset and a reproducible methodology for multimodal agricultural monitoring, with potential applications in visual verification, reduced reliance on costly field inspections, and data-driven agricultural policy implementation.

}


\maketitle

\section{Introduction}\label{Introduction}

\begin{figure*}[!t]
\centering
\includegraphics[width=\textwidth]{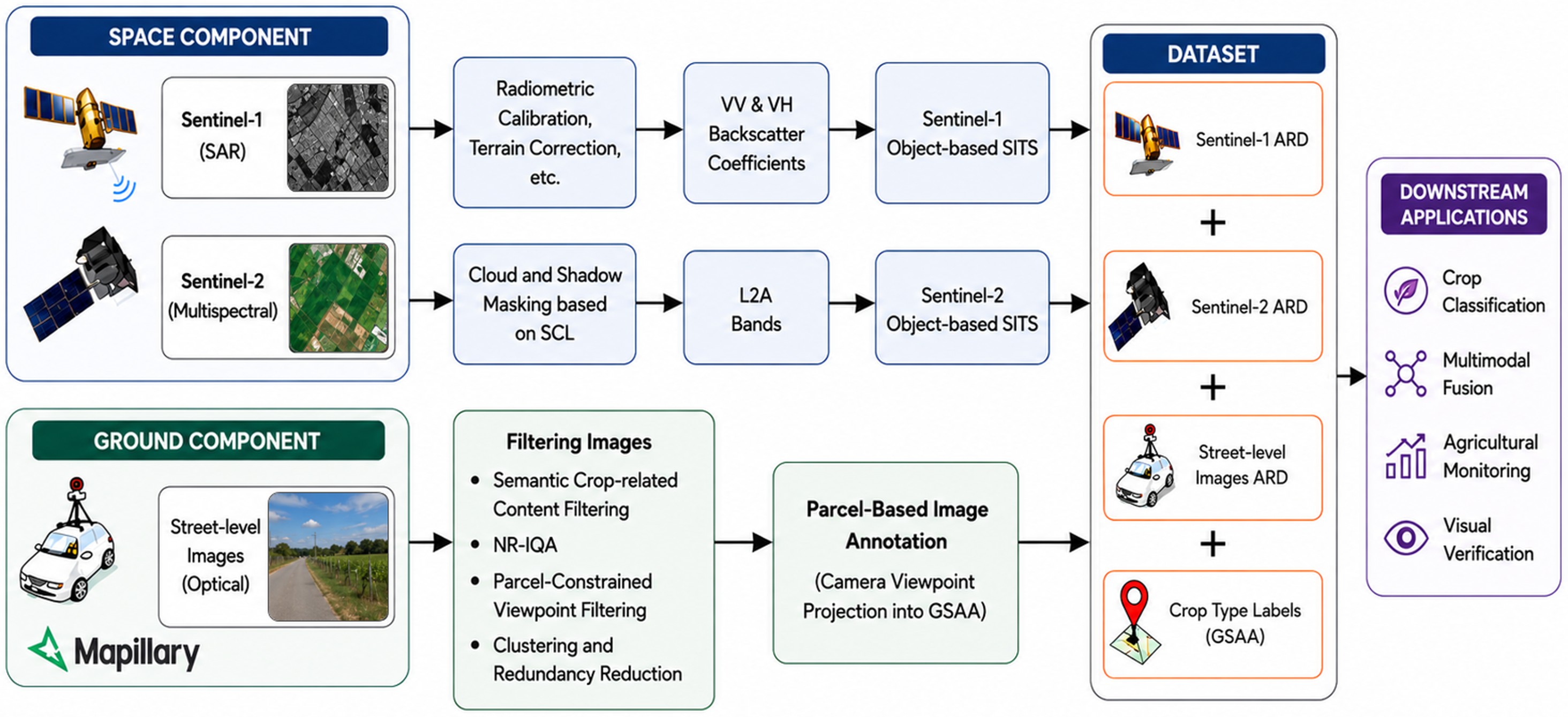}
\caption{Pipeline overview for constructing the Space2Ground 2.0 multi-source agricultural dataset. The framework integrates Sentinel-1 SAR, Sentinel-2 multispectral imagery, and crowdsourced Mapillary street-level images through automated preprocessing and parcel-level annotation, resulting in an analysis-ready dataset for crop monitoring and machine learning applications.}
\label{fig:image_annotation}
\end{figure*}

% Accurate and scalable parcel-level agricultural monitoring remains a significant challenge, particularly when relying solely on satellite Earth Observation (EO) data. 
The increasing availability of satellite Earth Observation (EO) data has transformed agricultural monitoring by enabling systematic observations over large geographic areas.
Openly accessible satellite missions (e.g., Copernicus) provide systematic observations with high temporal frequency and wide spatial coverage, enabling data-driven large-scale monitoring and early warning applications. Sentinel-2 optical and Sentinel-1 Synthetic Aperture Radar (SAR) constellations provide complementary observations, with optical sensors capturing vegetation-related spectral properties and SAR sensors offering structural and moisture-related information \cite{veloso-2017}. 
% However, optical data are affected by cloud cover,and require daylight conditions, while SAR signals operate in all weather and at night, but are more complex and less directly linked to crop characteristics. 
However, optical observations are affected by cloud cover and atmospheric conditions and require daylight, whereas SAR operates under all-weather and day–night conditions but is more complex to interpret and less directly linked to crop characteristics.
Moreover, the spatial resolution of these systems is not always sufficient to support accurate parcel-level decision-making, particularly for small or heterogeneous fields. Alternative platforms, such as very high resolution (VHR) commercial satellites or aerial imagery, provide finer spatial detail, but their higher cost, limited revisit frequency, or restricted coverage constrain large-scale operational use \cite{sishodia-2020}.

% To address these limitations, accurate and timely in-situ observations are required to complement remotely sensed data \cite{weiss-2020}. Traditional field surveys are costly, labor-intensive, and unable to provide a continuous and scalable flow of ground-truth information necessary for calibration, verification, and near real-time monitoring. In this context, crowdsourced street-level imagery emerges as a promising alternative, offering a cost-effective and scalable source of in-situ observations \cite{see-2016,dandrimond-2018-1}.
To address these limitations, accurate and timely in-situ observations are required to complement remotely sensed data \cite{weiss-2020}. Traditional field surveys are costly and labor-intensive, and are typically confined to specific areas or targeted inspection campaigns. As a result, they cannot provide the continuous and scalable flow of ground-truth information needed for calibration, verification, and near real-time monitoring. 
% In this context, crowdsourced street-level imagery emerges as a promising alternative, offering a cost-effective and scalable source of in-situ observations \cite{see-2016,dandrimond-2018-1}.
In this context, crowdsourced street-level imagery has emerged as a promising alternative, offering a cost-effective and scalable source of in-situ observations.
% Unlike satellite imagery, it provides a human-like, ground-level perspective of agricultural environments, capturing detailed visual characteristics of crops and their surrounding context that are difficult to observe from overhead satellite imagery \cite{see-2016,dandrimond-2018-1}.
Unlike top-down EO imagery, it provides a human-like, ground-level perspective of agricultural environments, capturing detailed visual characteristics of crops and local contextual information \cite{see-2016,dandrimond-2018-1}.

Recent years have seen increasing interest in the use of ground-level imagery for agricultural applications, including crop type identification, phenology monitoring, field boundaries detection, and visual verification of agricultural practices \cite{ringland-2019,dandrimont-2022,he-2022}. In some cases, ground observations have also been combined with satellite EO data to improve classification performance \cite{dandrimont-2018-2,choumos-2022}. 
However, the operational integration of ground-level imagery into scalable agricultural monitoring workflows remains limited. 
% However, most existing approaches remain limited to controlled field campaigns, manually curated datasets, or localized study areas, restricting their scalability and operational applicability. 
Opportunistic crowdsourced imagery introduces challenges such as heterogeneous acquisition conditions, sparse coverage, noisy or irrelevant content, and the absence of direct annotations \cite{karagiannopoulou-2022,huang-2024}. Beyond multimodal fusion itself, transforming such weakly structured imagery into reliable, parcel-linked, analysis-ready data remains a major practical challenge \cite{dellacqua-2017}.
These requirements become particularly relevant within evolving agricultural policy frameworks such as the Common Agricultural Policy (CAP), which increasingly rely on data-driven and performance-based monitoring mechanisms. The transition to the Area Monitoring System (AMS) requires continuous, large-scale observation of agricultural activities supported by robust and verifiable evidence. In this context, integrating spaceborne and ground-level observations can enhance monitoring accuracy, support desk on-the-spot-checks (OTSC), and facilitate dispute resolution processes \cite{sitokonstantinou-2022}.

Building upon the concept of Space-to-Ground data availability \cite{choumos-2022}, this paper presents Space2Ground 2.0, a framework for integrating satellite EO time series with street-level imagery for parcel-level agricultural monitoring. The proposed work addresses the gap between the growing availability of heterogeneous remote sensing data and their practical use in scalable agricultural monitoring systems.
The main contributions of this work are summarized as follows:
\begin{itemize}
\item An automated end-to-end methodology for transforming crowdsourced ground-level imagery into analysis-ready, parcel-linked data through quality control, semantic filtering, and spatial association.
\item A scalable multi-source framework for integrating satellite EO and street-level imagery for parcel-level agricultural monitoring, visual verification, and dispute resolution workflows.
% \item An openly available multi-source benchmark dataset supporting reproducible research and downstream AI applications, including crop classification.
\item An openly available multi-source benchmark dataset together with baseline ML/DL evaluations for multimodal crop classification, supporting reproducible research and downstream AI applications.
\end{itemize}

\section{Study Area \& Data Sources}\label{Materials}

\subsection{Cyprus Reference Parcel Data}

The study took place in Cyprus, a diverse Mediterranean agricultural environment characterized by mixed crop systems and relatively small parcel sizes (approximately 0.4 hectares). 
% Furthermore, the relatively high rate of erroneous farmer declarations (exceeding 10\%) makes it a particularly challenging setting for accurate agricultural monitoring.
% Furthermore, inaccuracies in farmer declarations and parcel delineations introduce additional uncertainty into the reference data, making accurate agricultural monitoring more challenging.
Reference parcel geometries and crop type labels were obtained from the Cyprus national Geospatial Aid Application (GSAA) system, managed by the Cyprus Agricultural Payments Organization (CAPO). GSAA serves as the official geospatial reference database for agricultural subsidy administration, compliance monitoring, and land parcel management under the CAP.
The GSAA dataset consists of geo-referenced parcel polygons accompanied by farmer-declared crop type information, which is used in this study as the reference label for assigning parcel-level labels to associated ground-level imagery and for downstream crop classification experiments.
The presence of erroneous farmer declarations (exceeding 10\% according to CAPO) introduces uncertainty into the reference data, making it a particularly challenging setting for accurate agricultural monitoring.
The national dataset for the 2022 growing season contains 325,673 agricultural parcels. Cereals dominate the agricultural landscape (47\%), followed by tree crops (20\%) and fallow land (14\%), while vineyards, permanent grasslands, potatoes, vegetables, vicia, and legumes account for the remaining area.
% The national dataset corresponding to the 2022 growing season contains 325,673 agricultural parcel geometries. More specifically, the agricultural landscape is dominated by cereal crops (approximately 47\%), followed by tree crops (20\%) and fallow land (14\%). Additional crop categories, including vineyards, permanent grasslands, potatoes, vegetables, vicia crops, and legumes, are present in smaller proportions. 

% Parcel sizes are generally small, with an average area of approximately 0.4 hectares, which increases the difficulty of accurate parcel-level monitoring using medium-resolution satellite imagery.

\subsection{Sentinel Data}

Copernicus Sentinel-1 and Sentinel-2 multi-temporal imagery were used to construct the space component of the proposed framework over agricultural parcels in Cyprus for the 2022 growing season.
Sentinel-2 Level-2A multispectral products were used to characterize vegetation dynamics through spectral information, while cloud and shadow contamination were masked using the Scene Classification Layer (SCL).
Sentinel-1 Level-1 Ground Range Detected (GRD) products acquired in Interferometric Wide (IW) swath mode were used to extract SAR backscatter information. Both ascending and descending acquisitions were included to improve temporal coverage and observation consistency. Preprocessing included orbit correction, radiometric calibration, terrain correction, and conversion of VV and VH backscatter coefficients into decibel (dB) scale.
Following preprocessing, satellite observations were aggregated at the parcel level using the GSAA geometries by computing mean band/polarization values within each parcel and acquisition date, and subsequently transformed into time series representations for downstream analysis.

% \subsection{Street-Level Data -- Cyprus Field Campaign}
\subsection{Street-Level Image Collection}

% \subsubsection{Mapillary as a Street-Level Data Source}\hfill\\[0.1cm]
% \subsubsection{Cyprus Field Campaign}\hfill\\[0.1cm]

\begin{figure}[ht!]
\begin{center}
		\includegraphics[width=1.0\columnwidth]{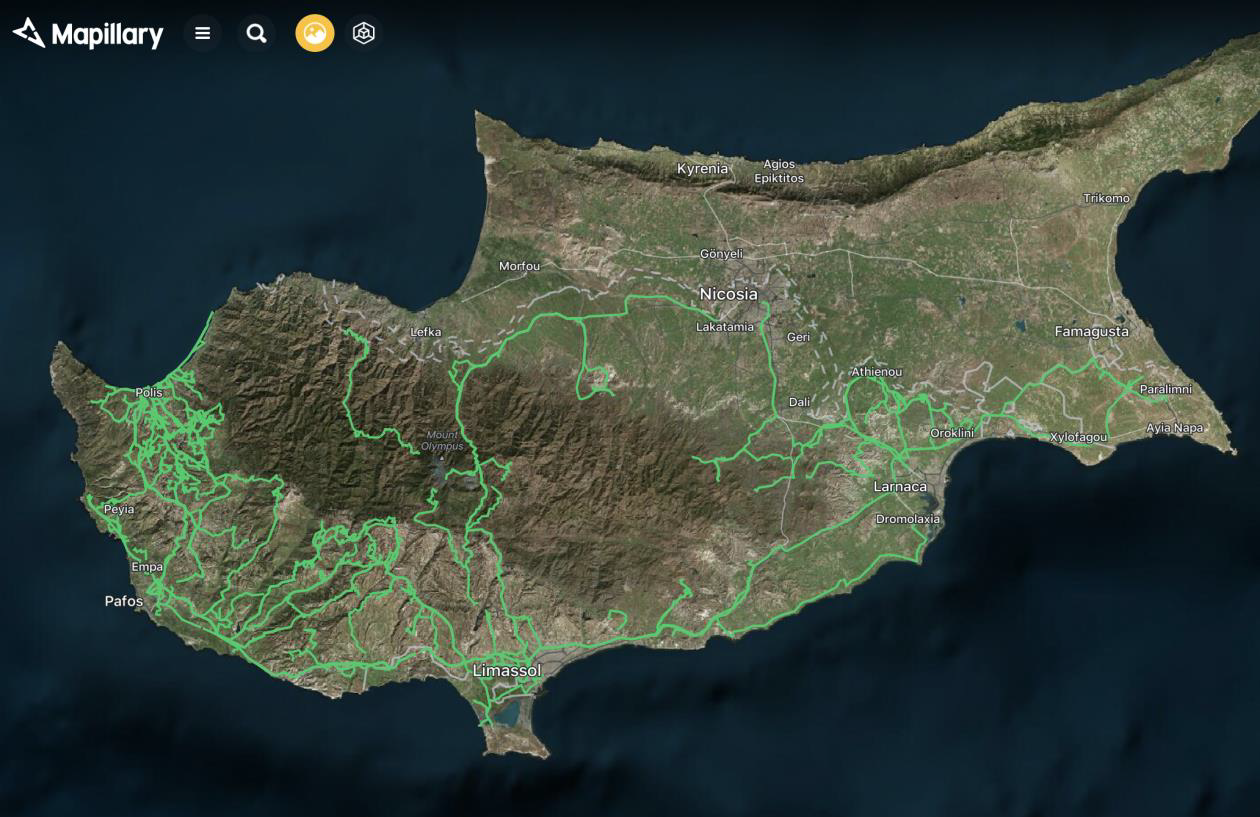}
	\caption{Mapillary street-level image acquisition coverage across Cyprus during the 2022 growing season. Green lines indicate the trajectories from which street-level observations were collected.}
\label{fig:mapillary_data}
\end{center}
\end{figure}

The ground component of the proposed framework is based on a dedicated street-level image acquisition campaign conducted across Cyprus throughout the 2022 growing season. The campaign was conducted by a single operator through repeated driving sessions distributed throughout the year, following road networks adjacent to main agricultural territories and, where accessible, field roads within agricultural areas.
A vehicle-based acquisition setup was designed using two side-mounted imaging devices to simultaneously capture left- and right-facing roadside views. Specifically, a GoPro HERO9 Black action camera (left side) and a Samsung Galaxy SM-A326B/DS smartphone (right side) were mounted on opposite sides of the vehicle, substantially increasing agricultural scene coverage and improving parcel visibility from both viewing directions.
% The campaign covered more than 10,000 km of road network across Cyprus (see Figure \ref{fig:mapillary_data}) and resulted in an initial collection exceeding 900,000 geo-tagged street-level images. Smartphone imagery was automatically uploaded to the Mapillary platform during acquisition, while GoPro imagery was transferred through SD card extraction and uploaded manually using batch ingestion workflows.
The campaign covered more than 10,000 km of road network across Cyprus (see Figure \ref{fig:mapillary_data}) and resulted in an initial collection exceeding 900,000 geo-tagged street-level images. Smartphone imagery was uploaded automatically to the Mapillary platform through its mobile application, while GoPro imagery was transferred via SD card extraction and subsequently uploaded using the Mapillary Desktop Uploader.
Mapillary was selected as the underlying platform due to its practical integration capabilities and built-in preprocessing tools. Beyond scalable API-based access to geo-tagged imagery, the platform provides standardized metadata (e.g., geolocation, timestamps, camera orientation), automated privacy-preserving processing such as face and license plate blurring, and image organization services that facilitate large-scale data management and downstream integration within automated geospatial workflows.

\section{Methodology \& Results}\label{Methods}

% \subsection{Mapillary Data Retrieval}

% Street-level imagery used for dataset generation was managed through the Mapillary platform. Image retrieval and metadata extraction were performed using a set of custom pipelines developed for the Mapillary API v4, which are publicly available\footnote{\url{https://github.com/gchoumos/mapillary}}\footnote{\url{https://github.com/Agri-Hub/Space2Ground-2.0}}. Besides image access, the API provides rich metadata and computer vision outputs, including object detections and semantic segmentations.

% The developed pipelines provide the following core functionalities:

% \begin{enumerate}
%     \item \textit{Image discovery and retrieval}: Retrieval of image sequences from specific users or organizations, extraction of associated image identifiers, and downloading of imagery at configurable quality levels.
    
%     \item \textit{Metadata extraction}: Acquisition of image metadata, including timestamps, geographic coordinates, camera orientation, sequence information, and other geospatial attributes required for downstream processing and parcel association.
    
%     \item \textit{Computer vision products}: Retrieval and visualization of object detections and semantic segmentation outputs provided by the Mapillary platform, enabling further content analysis and quality assessment.
% \end{enumerate}

% The ground component initially consists of 907,177 geo-referenced images, which are processed through automated quality-control pipelines to isolate vegetation regions and remove noise.

\subsection{Mapillary Data Retrieval}

Street-level imagery was acquired through the Mapillary platform, which provides public access to crowdsourced geo-referenced images together with rich image metadata and computer vision products. Image retrieval and metadata extraction were performed through a set of custom pipelines developed for the Mapillary API v4, which are publicly available\footnote{\url{https://github.com/gchoumos/mapillary}}\footnote{\url{https://github.com/Agri-Hub/Space2Ground-2.0}}. Besides image access, the API exposes a wide range of information, including acquisition timestamps, geographic coordinates, camera orientation, sequence identifiers, object detections, and semantic segmentation outputs. The developed pipelines automate large-scale street-level data acquisition and provide the following core functionalities:

\begin{enumerate}
    \item \textit{Image discovery and retrieval}: Retrieval of street-level image sequences from specific users or organizations, extraction of image identifiers, and downloading of imagery at configurable quality levels.

    \item \textit{Metadata extraction}: Extraction of image metadata, including acquisition timestamps, geographic coordinates, camera orientation, sequence identifiers, and additional geospatial attributes required for viewpoint projection and parcel-level annotation.

    \item \textit{Computer vision products}: Retrieval and visualization of object detections and semantic segmentation outputs generated by the Mapillary platform, supporting semantic content analysis and image quality assessment. 
\end{enumerate}

\begin{figure}[ht!]
\begin{center}
		\includegraphics[width=1.0\columnwidth]{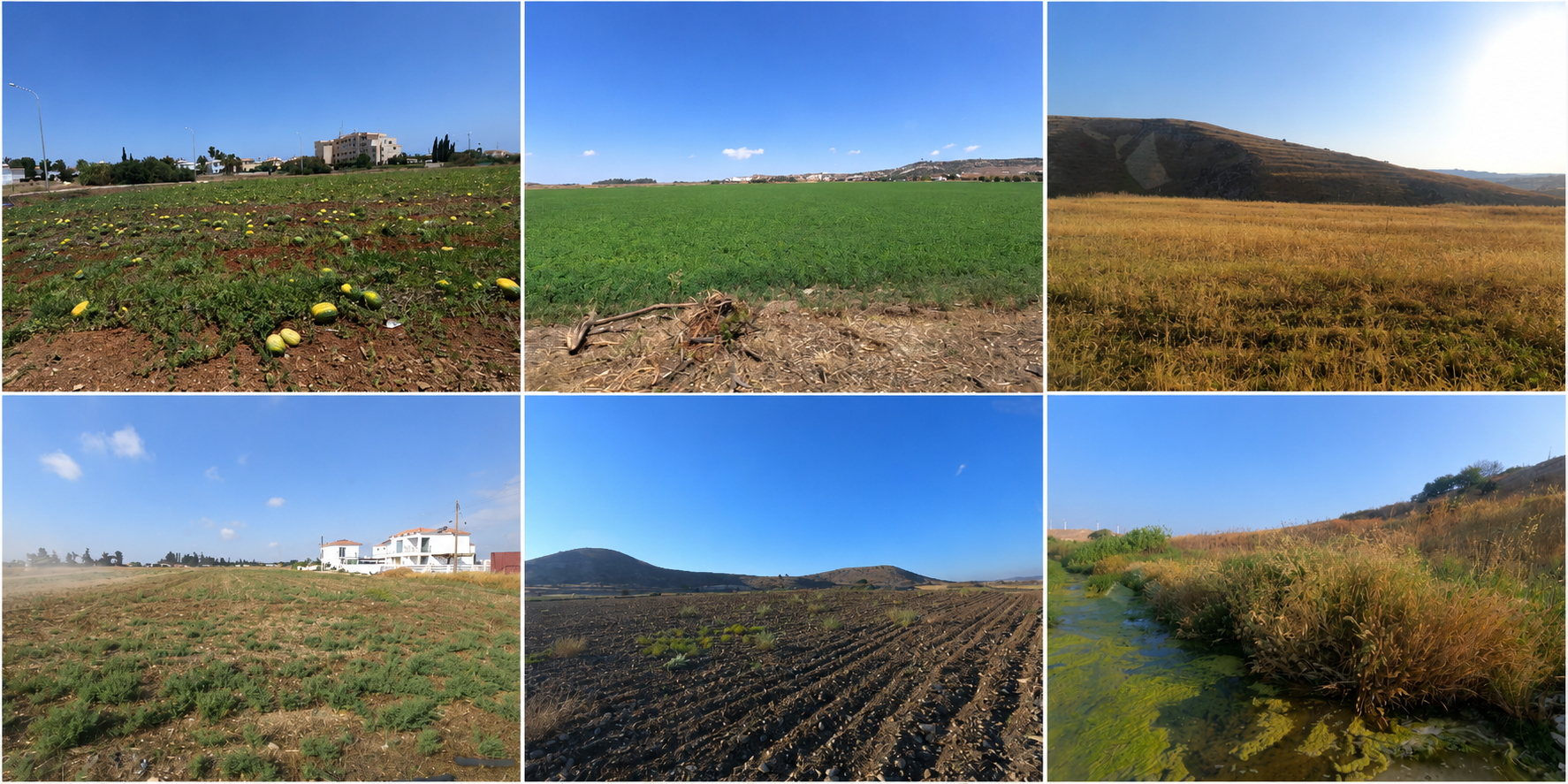}
	\caption{Representative examples of crowdsourced street-level agricultural images acquired and used for the construction of the Space2Ground 2.0 dataset.}
\label{fig:mapillary_samples}
\end{center}
\end{figure}

The ground component initially consists of 907,177 geo-referenced images, which are processed through automated quality-control pipelines to isolate vegetation regions and remove noise. Figure \ref{fig:mapillary_samples} shows some representative examples from the initial street-level image collection.

\subsubsection{Semantic Crop-Related Content Filtering}
\noindent

\begin{figure}[ht!]
\centering

\includegraphics[width=0.9\columnwidth]{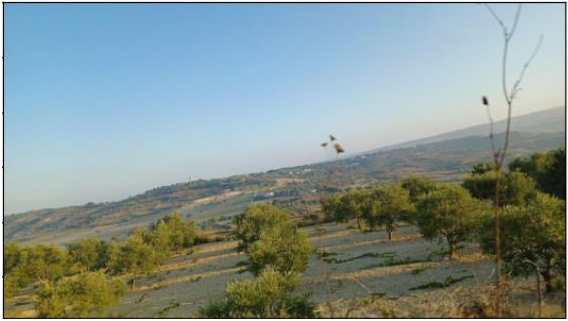}

\smallskip
(a) Original street-level image.

\vspace{0.2cm}

\includegraphics[width=0.9\columnwidth]{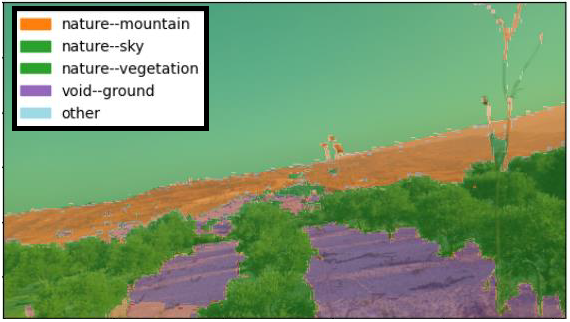}

\smallskip
(b) Semantic segmentation masks.

\caption{Example of street-level imagery and the corresponding semantic segmentation output provided by Mapillary.}
\label{fig:mapillary_segments}
\end{figure}

The raw street-level imagery collection contains a substantial amount of non-agricultural content, including roads, vehicles, buildings, and urban infrastructure. To increase dataset relevance, an initial semantic filtering stage was applied using the object detections and semantic segmentation products provided by the Mapillary API.
For each image, segmentation geometries corresponding to vegetation and terrain/soil -related categories were retrieved and decoded into image-space polygons. The cumulative area covered by these polygons was then calculated and expressed as a percentage of the total image extent. Images in which vegetation and soil occupied less than 20\% of the visible scene were discarded, removing predominantly non-agricultural content while retaining agriculturally relevant observations. Figure \ref{fig:mapillary_segments} presents an example of the segmentation outputs. This filtering stage reduced the dataset from 907,177 to 505,904 geo-referenced images, while more than 10 million segments were retained for subsequent processing.

\subsubsection{No-Reference Image Quality Assessment (NR-IQA)}
\noindent

Following semantic filtering, a no-reference image quality assessment (NR-IQA) stage was applied to remove low-quality images. Four state-of-the-art deep learning-based models, namely MANIQA \cite{MANIQA}, HyperIQA \cite{HyperIQA}, CLIP-IQA \cite{CLIPIQA}, and TReS \cite{TRES}, were used to independently assess image quality. For each NR-IQA model, images falling within the lowest 5\% of quality scores were considered low quality. An image was discarded if it was classified as low quality by at least three of the four models. This process removed 33,256 images, reducing the dataset from 505,904 to 472,648 images and improving the overall quality and consistency of the remaining collection.

\begin{figure*}[!t]
\centering
\includegraphics[width=\textwidth]{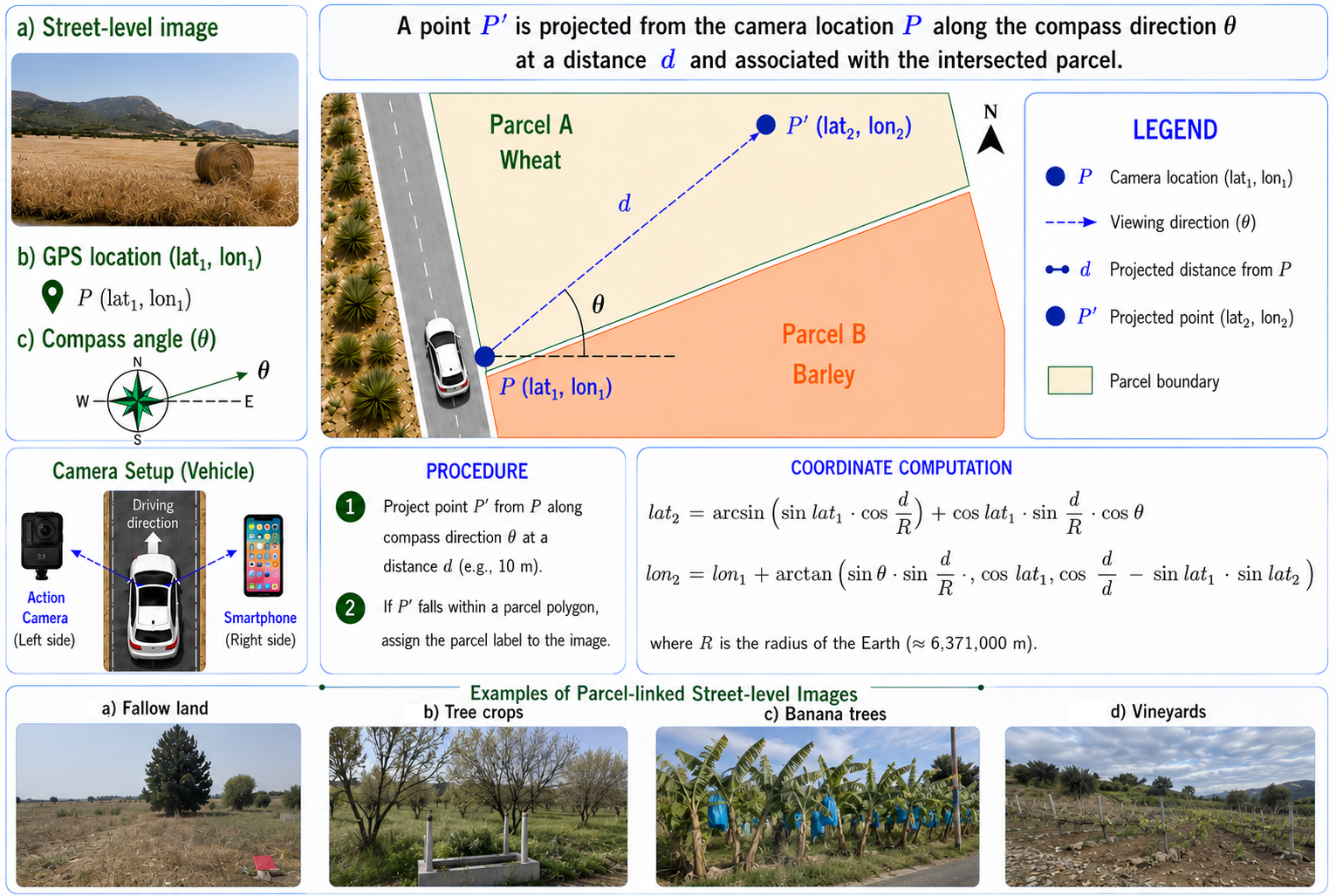}
\caption{Overview of the Space2ground 2.0 street-level image annotation workflow. A projected viewpoint is computed from the camera location and compass angle to associate images with agricultural parcels. Example parcel-linked images from different crop classes are also shown.}
\label{fig:image_annotation}
\end{figure*}

\subsubsection{Parcel-Level Annotation and Dataset Refinement}
\noindent

% Finally, the remaining images were spatially associated with agricultural parcels using their geo-referencing information and the GSAA parcel geometries. The annotation procedure follows the methodology proposed in \cite{sitokonstantinou-2022}, which exploits image location and viewing direction to identify the parcel observed in each street-level image.
% Unlike the original Space2Ground dataset, where front-facing cameras required image partitioning and separate viewpoint handling, the acquisition campaign in this study employed side-mounted cameras. Consequently, each image directly captured the adjacent agricultural landscape, simplifying the parcel association process.

The remaining images were subsequently associated with agricultural parcels using their geolocation, camera orientation, and the GSAA parcel geometries. The annotation procedure follows the viewpoint projection methodology proposed in \cite{sitokonstantinou-2022}, which exploits the camera position and viewing direction to identify the observed parcel. Unlike the original Space-to-Ground dataset, where front-facing cameras required image partitioning and multiple viewpoint estimations, the acquisition campaign in this study employed side-mounted cameras. Consequently, each image directly captured the adjacent agricultural landscape, simplifying the parcel association process and reducing viewpoint ambiguity.

For each image, the camera location was projected along its viewing direction to generate a reference point located 10 meters ahead of the camera position. Images whose projected viewpoint intersected an agricultural parcel inherited the corresponding crop label (see Figure \ref{fig:image_annotation}). When compass angle metadata were unavailable, frequently for images acquired and manually uploaded with the side-mounted GoPro camera, the viewing direction was reconstructed from the vehicle trajectory. Specifically, the vehicle heading was estimated by computing the azimuth between temporally consecutive geo-referenced image locations within the same Mapillary sequence. Camera-specific angular offsets (clockwise) were subsequently applied to account for the mounting configuration (+270° for the left-mounted GoPro camera and +90° for the right-mounted smartphone camera). The resulting viewing angle was then used for viewpoint projection and parcel association. This approach reduced ambiguities caused by road proximity, neighboring fields, and acquisition geometry, resulting in a collection of 75,303 parcel-linked street-level images.

Finally, an additional curation stage was applied to improve dataset consistency. Deep visual representations were extracted using a pre-trained VGG-16 network and subsequently reduced through Principal Component Analysis (PCA), retaining the 100 most informative components. The resulting feature vectors were grouped into 100 clusters using the k-means algorithm. Cluster contents were visually inspected, and clusters dominated by noisy, non-informative, or incorrectly associated images were discarded.
This refinement stage reduced the dataset to 46,050 annotated street-level images associated with 8,581 agricultural parcels.
The final curated collection comprises 14 crop classes (see Table \ref{tab:crop_distribution}), namely fallow land, vineyards, olive trees, barley, wheat, tree crops, potatoes, banana trees, permanent grasslands, vicia, oats, alfalfa, watermelons, and triticale, providing a benchmark dataset for downstream agricultural monitoring and machine learning applications.

\begin{table}[ht]
\centering
\caption{Number of street-level images remaining after each stage of the filtering pipeline.}
\label{tab:filtering_pipeline}
\resizebox{\columnwidth}{!}{%
\begin{tabular}{lr}
\toprule
\textbf{Filtering Step} & \textbf{Street-Level Images (\#)} \\
\midrule
Initial images & 907,177 \\
Semantic crop-related content filtering & 505,904 \\
No-reference image quality assessment & 472,648 \\
Parcel-constrained viewpoint filtering & 75,303 \\
Clustering and final dataset curation & 46,050 \\
\bottomrule
\end{tabular}
}
\end{table}

\begin{table}[ht]
\centering
\caption{Distribution of parcel instances and street-level image instances per crop class in the curated dataset.}
\label{tab:crop_distribution}
\resizebox{\columnwidth}{!}{%
\begin{tabular}{lrr}
\toprule
\textbf{Crop Class} & \textbf{Parcels (\#)} & \textbf{Street-Level Images (\#)} \\
\midrule
Fallow land & 1,688 & 8,652 \\
Vineyards & 1,261 & 7,479 \\
Olive trees & 1,292 & 6,687 \\
Barley & 1,258 & 6,541 \\
Wheat & 1,186 & 6,324 \\
Tree crops & 1,134 & 5,958 \\
Potatoes & 232 & 1,127 \\
Banana trees & 92 & 746 \\
Permanent grasslands & 121 & 744 \\
Vicia & 105 & 661 \\
Oats & 124 & 566 \\
Alfalfa & 24 & 261 \\
Watermelons & 35 & 164 \\
Triticale & 29 & 140 \\
\midrule
\textbf{Total} & \textbf{8,581} & \textbf{46,050} \\
\bottomrule
\end{tabular}
}
\end{table}

% \begin{figure*}[ht!]
% \centering
% \includegraphics[width=\textwidth]{figures/S2G_IVMSP_1-Space2Ground 2.0.drawio.png}
% \caption{Figure placement and numbering.}
% \label{fig:figure_placement}
% \end{figure*}

% \section{Benchmarking}\label{Methodology}

\subsection{Data Fusion for Monitoring and Classification}

% The integration of satellite and ground-level data enables complementary observations at different spatial and temporal scales.

% Satellite data provide consistent temporal coverage and large-scale patterns, while ground-level imagery captures fine-grained visual details and contextual information.

% Within this framework, data fusion is explored primarily as a means to:
% \begin{itemize}
%     \item Enhance crop classification performance
%     \item Support interpretation and validation of satellite-based predictions
%     \item Enable cross-modal consistency checks for monitoring purposes
% \end{itemize}

% Fusion strategies are evaluated to demonstrate the added value of integrating ground-level observations, rather than as the primary contribution of the work.

% The proposed framework reduces an initial collection of more than 900,000 crowdsourced images to a curated dataset of approximately 46,000 high-quality samples linked to agricultural parcels.

% This demonstrates the feasibility of transforming large-scale, unstructured data into analysis-ready datasets through automated processing.

% Experimental results show that:
% \begin{itemize}
%     \item Ground-level imagery provides complementary information to satellite data
%     \item Fusion approaches improve classification performance
%     \item The dataset supports both operational monitoring and machine learning benchmarking
% \end{itemize}

\begin{table*}[tb]
\centering
% \caption{Parcel-level crop classification performance of satellite-only, street-level-only, and multimodal fusion approaches. Results are reported as mean (± standard deviation) over five cross-validation folds. Overall accuracy is computed at parcel-level, while precision, recall, and F1-score correspond to macro-averaged scores across crop classes. For each fold, precision, recall, and F1-score are computed independently before averaging across folds. Bold values indicate the best-performing model within each modality group, while red values highlight the best overall result for each metric.}
\caption{Parcel-level crop classification performance of satellite-only, street-level-only, and multimodal fusion approaches. Results are reported as mean (± standard deviation) over five cross-validation folds. Overall accuracy is computed at the parcel level, while precision, recall, and F1-score are macro-averaged across crop classes. Bold values indicate the best result within each modality group, while red values highlight the best overall result for each metric.}
\label{tab:fusion_results}
\begin{tabular}{llcccc}
\toprule
\textbf{Modality} & \textbf{Model} & \textbf{Overall Accuracy (\%)} & \textbf{Precision (\%)} & \textbf{Recall (\%)} & \textbf{F1-Score (\%)} \\

\midrule
Satellite  
& Logistic Regression & 78.44 (±0.04) & 71.06 (±1.78) & 65.17 (±1.47) & 67.20 (±0.94) \\
& Random Forest & 77.02 (±0.25) & \textbf{78.11 (±2.98)} & 59.14 (±1.43) & 62.26 (±1.58) \\
& SVM & 78.60 (±0.70) & 74.37 (±2.13) & \textbf{66.11 (±2.13)} & \textbf{69.09 (±2.26)} \\
& XGBoost & \textbf{78.90 (±0.06)} & 74.77 (±0.82) & 66.08 (±1.13) & 68.92 (±0.84) \\
& GRU & 74.98 (±0.42) & 66.38 (±1.32) & 57.59 (±1.69) & 59.59 (±1.63) \\
& LSTM & 75.20 (±0.18) & 66.53 (±1.05) & 57.03 (±1.42) & 59.60 (±1.09) \\
& TempCNN & 75.62 (±0.32) & 66.96 (±1.49) & 59.18 (±1.16) & 61.27 (±0.79) \\
& TempCNN + LSTM & 75.77 (±0.06) & 67.43 (±1.65) & 59.75 (±0.76) & 62.23 (±0.89) \\

\midrule
Street-Level 
% & VIT-B-16 & 70.17 (±0.74) & 58.16 (±3.07) & 53.96 (±1.34) & 54.70 (±1.58) \\
% & VIT-B-32 & 69.25 (±0.93) & 58.17 (±1.14) & 52.69 (±1.61) & 54.15 (±1.70) \\
% & MOBILENET-V3-LARGE & 69.51 (±1.67) & 58.49 (±3.35) & 51.30 (±3.07) & 53.22 (±2.95) \\
% & SQUEEZENET1-1 & 68.67 (±1.04) & 57.91 (±3.08) & 50.20 (±1.83) & 51.94 (±2.09) \\
% & VGG16 & 68.50 (±1.09) & 55.34 (±2.62) & 51.50 (±2.64) & 51.81 (±2.51) \\
% & MOBILENET-V3-SMALL & 69.72 (±1.60) & 58.98 (±3.89) & 50.09 (±2.84) & 51.35 (±2.72) \\
% & EFFICIENTNET-B0 & 68.88 (±1.56) & 56.26 (±2.95) & 50.00 (±2.28) & 51.11 (±1.69) \\
% & VGG19 & 68.73 (±1.24) & 56.95 (±2.68) & 49.05 (±1.96) & 50.12 (±1.87) \\
% & DENSENET121 & 69.29 (±1.41) & 58.50 (±7.23) & 47.47 (±3.47) & 48.01 (±3.88) \\
% & RESNET18 & 67.40 (±1.62) & 53.82 (±6.81) & 46.05 (±1.11) & 46.64 (±1.83) \\
& VGG-16 & 68.50 (±1.09) & 55.34 (±2.62) & 51.50 (±2.64) & 51.81 (±2.51) \\
& VGG-19 & 68.73 (±1.24) & 56.95 (±2.68) & 49.05 (±1.96) & 50.12 (±1.87) \\
& ResNet-18 & 67.40 (±1.62) & 53.82 (±6.81) & 46.05 (±1.11) & 46.64 (±1.83) \\
& DenseNet-121 & 69.29 (±1.41) & 58.50 (±7.23) & 47.47 (±3.47) & 48.01 (±3.88) \\
& EfficientNet-B0 & 68.88 (±1.56) & 56.26 (±2.95) & 50.00 (±2.28) & 51.11 (±1.69) \\
& SqueezeNet-1.1 & 68.67 (±1.04) & 57.91 (±3.08) & 50.20 (±1.83) & 51.94 (±2.09) \\
& MobileNet-v3-small & 69.72 (±1.60) & \textbf{58.98 (±3.89)} & 50.09 (±2.84) & 51.35 (±2.72) \\
& MobileNet-v3-large & 69.51 (±1.67) & 58.49 (±3.35) & 51.30 (±3.07) & 53.22 (±2.95) \\
& ViT-B/16 & \textbf{70.17 (±0.74)} & 58.16 (±3.07) & \textbf{53.96 (±1.34)} & \textbf{54.70 (±1.58)} \\
& ViT-B/32 & 69.25 (±0.93) & 58.17 (±1.14) & 52.69 (±1.61) & 54.15 (±1.70) \\
\midrule
Early Fusion & VGG-16 + Logistic Regression & 81.35 (±0.72) & 79.84 (±1.48) & 78.62 (±1.35) & 79.10 (±1.21) \\
             & VGG-16 + Random Forest      & 81.82 (±0.64) & 80.56 (±1.32) & 79.28 (±1.26) & 79.78 (±1.08) \\
             & VGG-16 + SVM                & 82.21 (±0.58) & 81.04 (±1.24) & 79.92 (±1.18) & 80.36 (±1.01) \\
             & VGG-16 + XGBoost            & \textbf{82.65 (±0.52)} & \textbf{81.72 (±1.16)} & \textbf{80.43 (±1.09)} & \textbf{81.02 (±0.94)} \\
\midrule
Late Fusion  & XGBoost + ViT-B/16         & \textcolor{red}{\textbf{84.12 (±0.61)}} & \textcolor{red}{\textbf{83.35 (±1.08)}} & \textcolor{red}{\textbf{82.04 (±1.15)}} & \textcolor{red}{\textbf{82.78 (±0.96)}} \\
\bottomrule
\end{tabular}
\end{table*}

To evaluate the practical utility of the proposed dataset and framework, a series of crop classification experiments were conducted using satellite-only, street-level-only, and multimodal configurations. The objective was not to identify the optimal classification architecture, but rather to quantify the added value of integrating complementary spaceborne and ground-level observations.
For the satellite modality, parcel-level time series were constructed from Sentinel-1 (VV, VH) and Sentinel-2 spectral observations aggregated over the 2022 growing season. For the street-level modality, visual representations were extracted from the curated parcel-linked image collection using pre-trained Convolutional Neural Networks (CNNs) and Transformer-based image encoders. Image features belonging to the same parcel were subsequently aggregated through average feature pooling to obtain a single parcel-level representation. 
% Only parcels associated with both satellite observations and at least one street-level image were considered in the multimodal experiments, resulting in a subset of 8,581 agricultural parcels.
Two complementary fusion paradigms were investigated:

\begin{itemize}
\item \textbf{Early (feature-level) fusion}, where satellite and image-derived feature vectors were concatenated into a joint representation prior to model training.
% \item \textbf{Late (decision-level) fusion}, where independent satellite-based and image-based classifiers were trained separately and their prediction probabilities were combined through weighted averaging.
\item \textbf{Late (decision-level) fusion}, where independent satellite and street-level classifiers were trained separately and their prediction probabilities were combined through weighted averaging.
\end{itemize}

% Both conventional machine learning and deep learning approaches were evaluated. Machine learning models included Random Forest (RF), Support Vector Machines (SVM), Logistic Regression (LR), and Extreme Gradient Boosting (XGBoost). Deep learning baselines consisted of Long Short-Term Memory (LSTM), Gated Recurrent Units (GRU), and a lightweight temporal CNN architecture operating on the satellite time series. For multimodal deep learning experiments, image embeddings and temporal satellite representations were fused through a joint fully connected fusion layer.
% Performance was assessed using five-fold cross-validation at parcel-level. Table~\ref{tab:fusion_results} summarizes the overall classification accuracies obtained by the different approaches.
Both conventional machine learning and deep learning approaches were evaluated. For the satellite-only experiments, the conventional machine learning baselines included Logistic Regression (LR), Random Forest (RF), Support Vector Machines (SVM), and Extreme Gradient Boosting (XGBoost). The deep learning baselines included Gated Recurrent Units (GRU), Long Short-Term Memory (LSTM), a temporal convolutional neural network (TempCNN), and a hybrid TempCNN--LSTM architecture operating on the satellite time series.
For the street-level experiments, several pre-trained convolutional and Transformer-based image encoders were evaluated, including VGG, ResNet, DenseNet, EfficientNet, SqueezeNet, MobileNet, and Vision Transformer architectures. Ground-level predictions were aggregated at the parcel level to obtain a single prediction for each agricultural parcel.
For early fusion, street-level VGG-16 embeddings were concatenated with the satellite time series features and classified using LR, RF, SVM, and XGBoost. For late fusion, independent predictions from the satellite-based XGBoost model and the street-level ViT-B/16 model were combined through weighted averaging of their class probabilities.
To ensure direct comparability, all satellite-only, street-level-only, and multimodal experiments were evaluated using the same multimodal parcel subset and identical five-fold parcel-level cross-validation splits. 
% Overall accuracy and macro-averaged precision, recall, and F1-score were computed independently for each fold and subsequently reported as mean $\pm$ standard deviation. 
Table~\ref{tab:fusion_results} summarizes the resulting classification performance.

The results show that satellite observations provide the strongest single-modality performance. Among the satellite-only models, XGBoost achieved the highest overall accuracy, reaching 78.90\%, while SVM obtained the highest F1-score among the single-modality satellite baselines, with 69.09\%. This confirms the importance of Sentinel-1 and Sentinel-2 time series information for crop type classification, as satellite observations capture seasonal vegetation dynamics and crop phenological patterns throughout the growing season.
Street-level imagery achieved lower standalone performance than satellite observations, with ViT-B/16 obtaining the best street-level results in terms of both overall accuracy and F1-score, reaching 70.17\% and 54.70\%, respectively. This lower performance is expected, since street-level observations are opportunistic and usually consist of one or a limited number of images per parcel, limiting their ability to represent crop temporal evolution. Nevertheless, ground-level imagery captures fine-scale visual characteristics such as crop structure, canopy appearance, planting patterns, and local management conditions, which are not directly observable from medium-resolution satellite imagery.

The benefit of combining the two modalities is clearly demonstrated by the multimodal experiments. The best early-fusion configuration, VGG-16 + XGBoost, achieved 82.65\% overall accuracy and 81.02\% F1-score. This corresponds to an improvement of approximately 4\% in overall accuracy compared with the best satellite-only model, and a substantial increase in F1-score, indicating that the integration of street-level information improves class-balanced performance rather than only overall accuracy.
Late fusion achieved the best overall performance. The combination of XGBoost satellite predictions with ViT-B/16 street-level predictions reached 84.12\% overall accuracy, 83.35\% precision, 82.04\% recall, and 82.78\% F1-score. Compared with the best satellite-only model, late fusion improved overall accuracy by approximately 5\%. The improvement is even more pronounced for F1-score, highlighting that multimodal fusion improves performance across crop classes and not only for the dominant classes.

Overall, the results indicate that satellite and street-level observations provide complementary information for parcel-level crop classification. While satellite time series remain the most informative individual data source, street-level imagery contributes additional visual evidence that improves classification when integrated with satellite observations. The stronger performance of late fusion compared with early fusion suggests that the two modalities have different statistical characteristics and are more effectively combined after each modality has learned its own decision representation. This supports the use of multimodal fusion modalities in agricultural monitoring workflows, particularly when complementary spaceborne and ground-level observations are available.

\section{Discussion}\label{Discussion}

% discuss limitations such as images from the side of the road, physical obstacles, or false farmers' declarations (especially for fallow land). Give visual examples. Note errors in the cameras' geo-locations again.
% Discuss on manually upload, which gives the oppportunity of a quick manual filtering. Discuss on labels using farmers declarations (portion of errors)
% Discuss on labeling using GPT ...find relative papers existing

\begin{figure}[ht!]
\begin{center}
		\includegraphics[width=1.0\columnwidth]{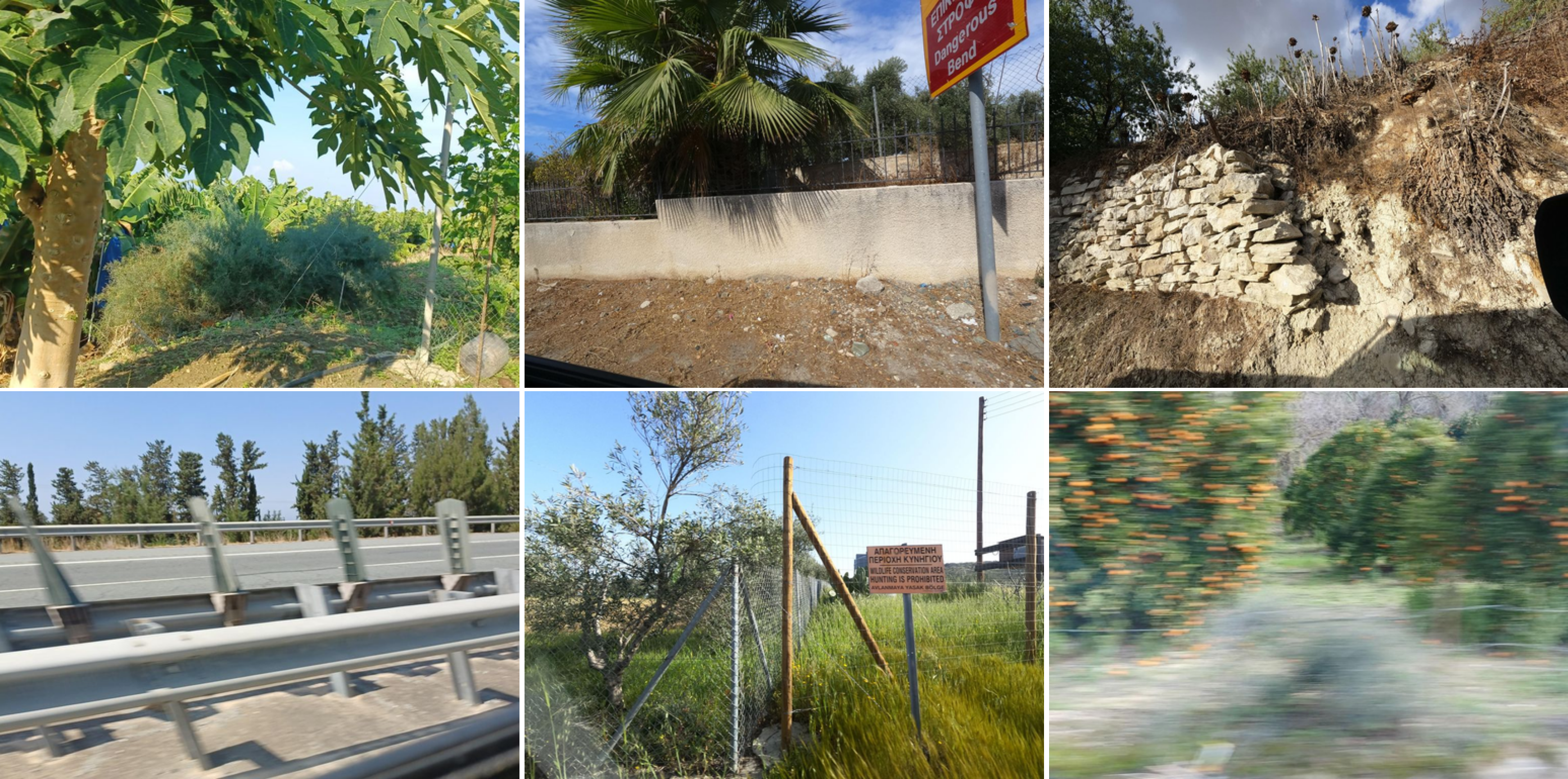}
	\caption{Representative examples of challenges encountered during parcel-level annotation of street-level imagery. The main sources of noise arise from physical occlusions by dense vegetation or fences, large road-to-parcel distances, images acquired near parcel boundaries where ambiguous visual information may be captured within the same field of view, and motion blur caused by high vehicle speeds.}
\label{fig:limitations}
\end{center}
\end{figure}

Although the proposed framework provides an automated and scalable methodology for constructing parcel-linked street-level agricultural datasets, several limitations remain. First, the acquisition process is inherently constrained by road accessibility, as observations are primarily collected from public road networks. Consequently, parcels located far from roads or surrounded by dense vegetation may not be adequately represented. Furthermore, the distance between the road and the target parcel may substantially reduce the visible portion of the crop, limiting the amount of useful visual information available for annotation. In addition, physical obstacles such as trees, fences or surrounding vegetation, parked vehicles, buildings, or terrain variations can partially or completely occlude the target parcel, resulting in incomplete or misleading visual observations. 
Ambiguities may also arise when images are acquired near the boundaries of adjacent parcels, where the camera viewing direction may include multiple crop types within the same field of view despite the image being assigned a single parcel label.
Figure \ref{fig:limitations} presents representative examples of such cases. Another important source of uncertainty originates from the geo-localization accuracy of consumer-grade imaging devices. Although viewpoint projection substantially improves parcel association by exploiting camera orientation, positioning errors introduced by GPS inaccuracies or compass estimation may still project observations onto neighboring parcels, particularly in regions characterized by relatively small agricultural parcels or fragmented field patterns.
These errors may be exacerbated when the vehicle is moving at higher speed, or when camera orientation metadata is unavailable and must be reconstructed from the acquisition trajectory.
Collectively, these sources of uncertainty are likely reflected in the comparatively lower standalone classification performance of the ground component, highlighting the challenges associated with opportunistic crowdsourced imagery.

Beyond acquisition-related uncertainties, parcel annotations inherit the uncertainty associated with the official GSAA declarations. While these labels represent the operational reference used within the CAP, they are not error-free. In the case of Cyprus, declaration inconsistencies may arise from delayed updates, boundary inaccuracies, agronomic ambiguities (e.g., fallow land), data entry errors, and occasional intentional misreporting. These factors inevitably propagate label noise into the generated dataset. This issue is particularly evident for fallow land, where parcels declared as fallow may exhibit substantial spontaneous vegetation growth or even active cultivation during image acquisition, resulting in an apparent mismatch between the visual evidence and the assigned administrative label. Consequently, some samples that appear incorrectly labeled may actually reflect limitations of the reference data rather than shortcomings of the proposed annotation methodology.

The proposed dataset is also subject to the sampling biases inherent in crowdsourced imagery.
Agricultural regions with denser road networks or greater user activity are better represented than remote areas, while acquisitions are unevenly distributed throughout the growing season. These limitations are intrinsic to opportunistic image collection rather than specific to the proposed framework and should be considered when interpreting large-scale street-level datasets. 
In addition, the current dataset was generated from imagery contributed by only a single operator, constraining the number and diversity of observations. Contributions from more users could provide imagery from additional routes, viewpoints, and phenological stages.
Despite these challenges, the processing pipeline intentionally preserves opportunities for lightweight human intervention. Rather than requiring exhaustive manual inspection of individual images, the clustering-based refinement stage groups visually similar observations, enabling rapid identification of systematic errors, non-agricultural content, and incorrectly associated parcels. Furthermore, manually uploaded imagery through the Mapillary platform offers an extra quality-control mechanism, allowing contributors to review observations before publication and remove blurred images, incorrect viewpoints, or irrelevant scenes with minimal additional effort. These human-in-the-loop components provide a practical compromise between fully automated dataset generation and expensive manual annotation while maintaining scalability for large image collections.

From an operational perspective, the proposed methodology is largely independent of the study area. Since it relies primarily on openly available satellite observations, crowdsourced street-level imagery, and parcel boundaries, the framework can be readily transferred to other countries where similar data sources exist. 
In addition, it should be noted that the proposed sequence of processing steps represents one practical implementation of the framework rather than a fixed workflow. Depending on the characteristics of the available data, implementation requirements, and computational constraints, individual processing steps may be reordered, omitted, or complemented with additional modules to better suit specific applications. Consequently, different workflow configurations may lead to variations in processing time, and the composition of the resulting dataset.
Looking forward, further improvements in dataset quality could be achieved by reducing the dependence on administrative declarations through recent advances in weakly supervised learning, self-training, and vision-language foundation models. Rather than relying exclusively on farmer-reported labels, these approaches offer the potential to automatically infer semantic information directly from street-level imagery, enabling large-scale dataset refinement with minimal manual intervention. For example, a relatively recent work \cite{soler-2024} combined GPT-4V-generated zero-shot labeled street view images with satellite time series to produce large-scale crop maps. Hence, within the proposed Space2Ground framework, similar approaches could be employed to validate farmer declarations, identify potentially mislabeled parcels, and generate pseudo-labels for previously unlabeled observations. Combined with active learning approaches that prioritize only uncertain samples for manual verification, such methods could substantially reduce annotation effort while continuously improving dataset quality. More broadly, these developments have the potential to transform Space2Ground from a static benchmark dataset into a continuously evolving agricultural monitoring framework, where crowdsourced street-level imagery, satellite observations, and multimodal AI models collaboratively refine parcel annotations and support increasingly reliable operational monitoring under the CAP requirements.

\section{Conclusions}\label{Conclusions}

This paper presented Space2Ground 2.0, a scalable multi-source framework for integrating Sentinel-1, Sentinel-2, and crowdsourced street-level imagery into a unified parcel-level dataset for agricultural monitoring. The proposed pipeline automates the transformation of large volumes of opportunistic street-level images into analysis-ready data through semantic filtering, image quality assessment, viewpoint-based parcel annotation, and dataset refinement. Applied over Cyprus for the 2022 growing season, the framework resulted in an openly available benchmark dataset comprising 46,050 annotated street-level images linked to 8,581 agricultural parcels.
The practical value of the proposed dataset was demonstrated through multimodal crop classification experiments combining satellite time series and street-level observations. The results showed that ground-level imagery provides complementary information to satellite EO data, leading to consistent improvements over single-modality approaches, with late fusion achieving the highest overall classification performance. These findings highlight the benefits of integrating complementary spaceborne and ground observations for parcel-level agricultural monitoring.
Beyond crop classification, the proposed framework provides a reproducible methodology for generating parcel-linked street-level datasets that can support agricultural monitoring, visual verification, and dispute resolution, while serving as a proof of concept for extending the Space2Ground approach to a broader range of EO downstream applications.

\section*{Data Availability}

The Space2Ground 2.0 dataset, is publicly available through Zenodo at: \url{https://doi.org/10.5281/zenodo.21219542}.

\section*{Acknowledgements}\label{Acknowledgements}

This research has received funding from the E-SPFdigit project under the European Union's Horizon Europe Programme (Grant Agreement No. 101157922). Access to restricted Cyprus GSAA data was provided by the Cyprus Agricultural Payments Organization (CAPO) through the CALLISTO project (Grant Agreement No. 101004152). The data were made available exclusively for research purposes and did not include any personal, identifiable, or farmer-related information, ensuring full compliance with applicable data protection and privacy requirements. The authors would like to thank Mr. Nikos Daniil (CAPO) for his valuable contribution to the street-level image acquisition campaign and field data collection efforts across Cyprus.

{
	\begin{spacing}{1.17}
		\normalsize
		\bibliography{references} % Include your own bibliography (*.bib), style is given in isprs.cls
	\end{spacing}
}

\end{document}